\documentclass[letterpaper]{article} %
\usepackage[]{aaai23}  %
\usepackage{times}  %
\usepackage{helvet}  %
\usepackage{courier}  %
\usepackage[hyphens]{url}  %
\usepackage{graphicx} %
\usepackage{natbib}  %
\usepackage{caption} %
\usepackage{algorithm}
\usepackage{algorithmic}
\usepackage{booktabs}

\usepackage{newfloat}
\usepackage{listings}
\DeclareCaptionStyle{ruled}{labelfont=normalfont,labelsep=colon,strut=off} %
\floatstyle{ruled}
\newfloat{listing}{tb}{lst}{}
\floatname{listing}{Listing}
\usepackage{amsfonts}
\usepackage{bm}
\usepackage{subcaption}

\title{Federated Learning over Harmonized Data Silos}
\author {
    Dimitris Stripelis,
    Jos\'{e} Luis Ambite \\
}
\affiliations {
    Information Sciences Institute, University of Southern California, Los Angeles, CA, 90007 \\
    \{stripeli, ambite\}@isi.edu
}

\newcommand{\metis}{MetisFL}

\newcommand{\SystemDesignShort}{FLINT}
\newcommand{\SystemDesignLong}{Federated Learning and Integration}

\begin{document}

\maketitle

\begin{abstract}
Federated Learning is a distributed machine learning approach that enables geographically distributed data silos to collaboratively learn a joint machine learning model without sharing data. Most of the existing work operates on unstructured data, such as images or text, or on structured data assumed to be consistent across the different sites. 
However, sites often have different schemata, data formats, data values, and access patterns.
The field of data integration has developed many methods to address these challenges, including techniques for data exchange and query rewriting using declarative schema mappings, and for entity linkage. 
Therefore, we propose an architectural vision for an end-to-end \SystemDesignLong~system, incorporating the critical steps of data harmonization and data imputation, to spur further research on the intersection of data management information systems and machine learning.

\end{abstract}

\section{Introduction}

Federated Learning (FL)~\cite{mcmahan2017communication} and Analytics~\cite{fedanalytics} is a distributed learning approach that allows to collaboratively train machine learning and other statistical models from decentralized data. 
Since different sites often cannot share their data due to competitiveness, legal, and privacy constraints, Federated Learning keeps the data at its original source, and pushes the learning process, usually training a neural network, down to each source. A central server coordinates the distributed training among the participating data sources and aggregates the locally learned neural models (or other statistics) to compute a global model. Training can be performed under strong privacy guarantees through homomorphic encryption~\cite{rivest1978data}.
Federated learning~\cite{kairouz2019advances,li2020federated,yang2019federated} has been very effective both for edge and mobile devices~\cite{lim2020federated}, and for federations of business organizations~\cite{liu2020fedvision}, or biomedical research consortia~\cite{rieke2020future,kaissis2020secure}.

The same privacy constraints that prevent data sharing across sites also inherently create isolated data environments, i.e., data silos~\cite{jain2003out}. Most work in Federated Learning focuses on solving challenges related to the distributed learning optimization problem~\cite{li2020federated,wang2021field}, but the core challenge of data harmonization across silos is overlooked. Existing systems~\cite{li2021survey} assume that the local data at the participating sources (which are the input into the learning model) follow the same schema, format, semantics, and storage and access capabilities. Such an assumption does not hold in realistic learning scenarios, where geographically distributed data sources have their own unique data specifications; a challenge that is commonly observed in Federated Database Management Systems~\cite{heimbigner1985federated,sheth1990federated}, Data Integration~\cite{doan2012}, and Data Exchange~\cite{fagin2005}.
Therefore, we present an architecture for a~\SystemDesignLong, which includes data harmonization and data imputation as core components.

\vspace{-2mm} %
\begin{figure}[htpb]
  \centering
  \includegraphics[width=0.7\linewidth]{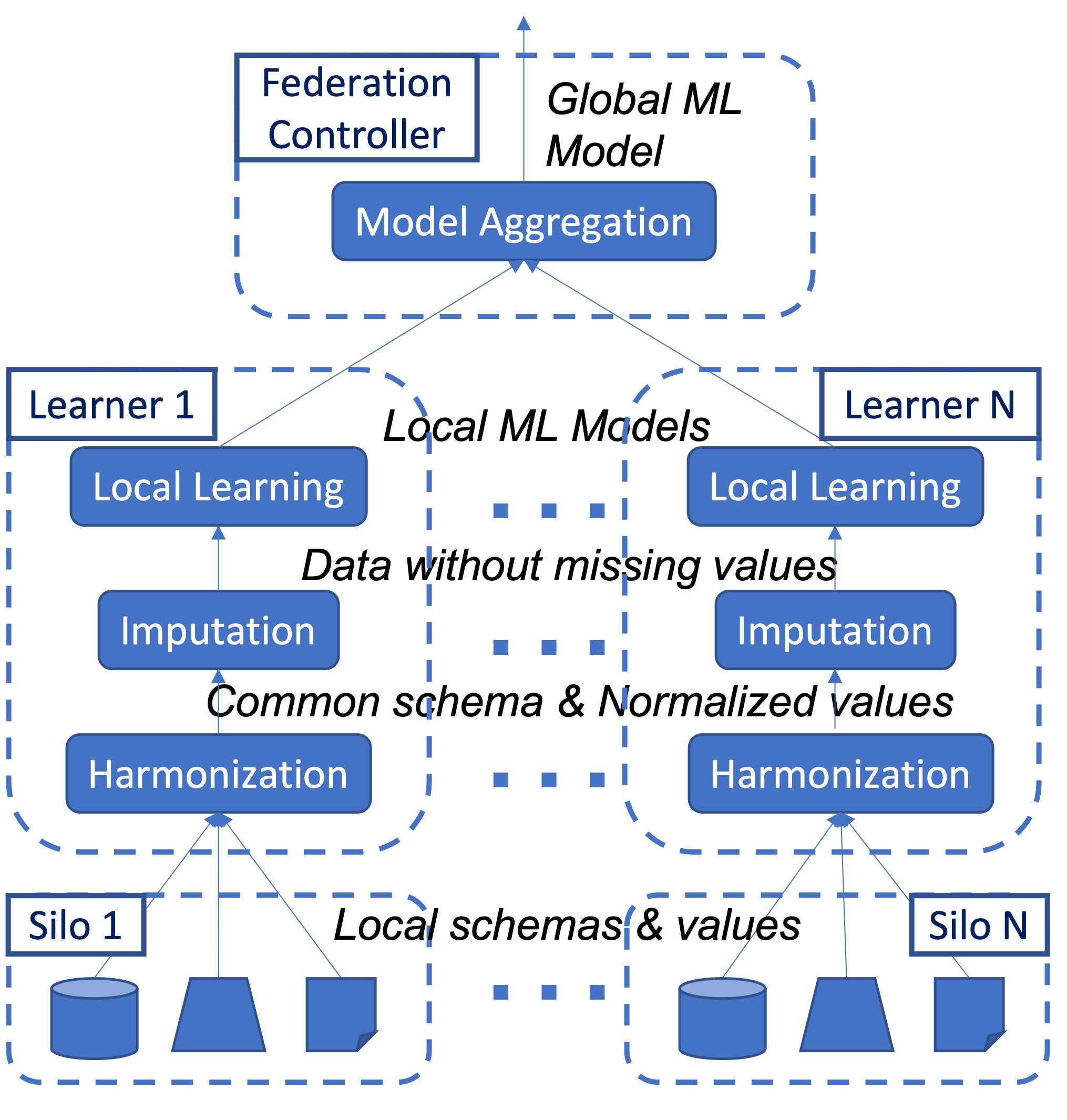}
  \vspace{-2mm}
  \caption{A Harmonized Federated Learning Workflow.}
  \label{fig:fldms-architecture}
\end{figure}
\vspace{-2mm} %

Figure~\ref{fig:fldms-architecture} shows our high-level architecture. 
The \textit{Data harmonization} component maps each local schema (and values) to a common schema (and values) agreed by the federation, which we advocate should be done through declarative schema 
mappings~\cite{halevy2001,fagin2005,gottlob2014s,doan2012,dong2015,xiao2018s}. 
This common schema intends to support multiple learning scenarios over the domain of the data. Since not all sources may have values for all the attributes in the common schema, it is often necessary to impute missing values to improve the precision of statistical studies~\cite{kose2020effect} and reduce prediction bias~\cite{ayilara2019impact}, specially in clinical studies.
This has implications for the data integration methods used: instead of removing answers with skolems/labelled nulls, these can be preserved and the missing values imputed.
Altogether, we identify the following core challenges that need to be solved to facilitate deployment of Federated Learning solutions in real-world settings:
\begin{itemize}
    \item Private and secure data harmonization and normalization across federated learning silos.
    \item Enable federated training over missing values using imputation to improve learning efficiency and reduce bias.
    \item Improve data query access patterns for efficient ingestion of training data into siloed learning models.
\end{itemize}
In the remainder, we provide the necessary background of Federated Learning optimization, the need for federated data integration and imputation, and our proposed architecture.

\section{Background}
\label{sec:background}

In a Federated Learning environment consisting of $\mathcal{N}$ participating learners, we want to minimize the objective function~\cite{mcmahan2017communication,wang2021field}:
\begin{equation}\label{eq:FederatedFunction}
    F(\bm{x}) = \mathbb{E}_{k \sim \mathcal{K}}[F_k(\bm{x})] \quad F_k(\bm{x}) = \mathbb{E}_{\xi \sim \mathcal{D}_k}[\ell_k(\bm{x}, \xi)]
\end{equation}
where $\bm{x}$ represents the model parameters and $\mathcal{K}$ the learner distribution selection over the population of $\mathcal{N}$ learners. Every learner computes its local objective by minimizing the empirical risk $F_k$ over its local data distribution $\mathcal{D}_k$, with $\mathcal{D}_i \neq \mathcal{D}_j, \forall i \neq j$ and $\ell_k(\bm{x},\xi)$ the local loss function. The global objective function $F(\bm{x})$ can take the form of an empirical risk minimization objective $F(\bm{x})=\sum_{k=1}^{P}\frac{p_k}{\mathcal{P}}F_k(w)$ with $p_k$ denoting the contribution value of learner $k$ in the federation and $\mathcal{P}=\sum p_k$ the normalization factor; hence $\sum_{k}^\mathcal{K} \frac{p_k}{\mathcal{P}}=1$. In the original Federated Average (FedAvg) algorithm~\cite{mcmahan2017communication}, the contribution value of every learner $k$ equals its local training dataset size, $p_k = \left|\mathcal{D}_k\right|$ and a central server aggregates local models updates at given synchronization points known as federation rounds.

Federated \textbf{optimization} methods need to address several challenges that do not usually arise in centralized settings, namely learners' communication constraints, local data and computational heterogeneity, learning topology, and security and privacy. To provide convergence guarantees in the presence of these learning constraints, existing methods decouple the federated optimization problem into global (server-side) and local (client-side). Server-side optimization, e.g., ~\cite{reddi2020adaptive}, refers to the algorithm applied while merging learners' local models, and client-side optimization, e.g., ~\cite{li2018federated}, to the algorithm applied during learners' local training. Recent surveys~\cite{kairouz2019advances,wang2021field} provide further details.

Federated Learning systems can be structured with different \textbf{topologies} depending on communication constraints~\cite{rieke2020future,kairouz2019advances}. 
In a \textit{centralized} (star) topology learners communicate through a central entity (controller). In a \textit{decentralized} topology learners communicate directly with each other (peer-to-peer) without any central coordinator.
Hierarchical and hybrid approaches~\cite{rieke2020future,bellavista2021decentralised}, where multiple sub-aggregators can co-exist~\cite{bonawitz2019towards}, have also been explored.

Federated Learning has been applied in two \textbf{environments} with complementary properties: \textit{cross-silo} FL consisting of tens or hundreds of reliable, stateful learners with ample computational resources (e.g., geo-distributed datacenters, hospital networks), and \textit{cross-device} FL consisting of thousands or millions of unreliable, stateless learners with limited computation and communication capacity (e.g., IoT sensors, cell-phones)~\cite{kairouz2019advances}.
For simplicity of exposition, we focus on a cross-silo centralized topology where all learners train the same neural network.

Federated training can follow different \textbf{communication protocols}. In a synchronous protocol~\cite{mcmahan2017communication}, each learner trains for a given number of epochs. The controller waits for all learners participating in the current round to finish their local training before computing a new global model, which may cause fast learners to remain idle while waiting for slow learners.
In an asynchronous protocol~\cite{xie2019asynchronous}, each learner trains for a given number of epochs and immediately sends its local model to the controller. There is no idle time, but the global model may be computed over stale local models.
In a semi-synchronous protocol~\cite{stripelis2022semi}, all learners train for the same time period before sharing their local models with the controller. This approach avoids idle time, but learners may perform different amounts of work. Empirically, this approach often performs better.

To improve \textbf{privacy and security} during federated training, a range of privacy-preserving federated learning algorithms have been recently proposed~\cite{kairouz2019advances} based on private-aware training and secure aggregation. To ensure protection against different attack scenarios, such as membership inference attacks~\cite{gupta2021membership} or collusion attacks~\cite{ma2021privacy}, a Federated Learning solution needs to incorporate both approaches. In the case of private training, learners can train the global model on their local data using (differential) private-aware methods, such as DP-SGD~\cite{abadi2016deep}. In the case of secure aggregation, the aggregation of local models by the controller can be performed through Secure Multi-Party Computation~\cite{bonawitz2016practical} or Fully Homomorphic Encryption (FHE)~\cite{zhang2020batchcrypt,stripelis2021secure} schemes. In the FHE setting~\cite{stripelis2021secure}, the controller sends the encrypted global model weights (ciphertext) to each learner, a trusted entity generates the public and private key pair and shares the key pair with every other learner and only the public key with the controller. The learners decrypt the encrypted weights using a private key, train the decrypted model on their local dataset, encrypt the new local model (ciphertext) using the public key and send it back to the controller. Upon receiving the encrypted local models, the controller performs a private weighted-aggregation step over the ciphertexts to compute the global model.

Depending on the distribution of the training records used to train the federated model across the participating learners, different \textbf{data partitioning} schemes have been investigated~\cite{yang2019federated,yin2021comprehensive}. Let $\mathcal{I}$ denote the id (entity) space, $\mathcal{X}$ the feature space and $\mathcal{Y}$ the label space of the training records, these schemes are:
\begin{itemize}
    \item \textit{Horizontal Federated Learning (HFL)}. Learners own data with the same feature and label space but different id space, e.g., participants in a research consortium with multiple sites.
    \begin{equation}
    \mathcal{X}_i = \mathcal{X}_j,\ \ \mathcal{Y}_i = \mathcal{Y}_j, \ \  I_i\neq I_j,\ \ \forall \mathcal{D}_i, \mathcal{D}_j, i\neq j
    \end{equation}
    \item \textit{Vertical Federated Learning (VFL)}. Learners own data from the same id space but with different feature and/or label space, e.g., same patients across different hospitals. 
    \begin{equation}
    \mathcal{X}_i \neq \mathcal{X}_j,\ \ \mathcal{Y}_i \neq \mathcal{Y}_j, \ \  I_i = I_j,\ \ \forall \mathcal{D}_i, \mathcal{D}_j, i\neq j
    \end{equation}
    \item \textit{Federated Transfer Learning (FTL)}. Learners own completely disjoint datasets, with different id, feature and label space, e.g., different customers across different organizations.
    \begin{equation}
    \mathcal{X}_i \neq \mathcal{X}_j,\ \ \mathcal{Y}_i \neq \mathcal{Y}_j, \ \  I_i\neq I_j,\ \ \forall \mathcal{D}_i, \mathcal{D}_j, i\neq j
    \end{equation}
\end{itemize}
Most federated learning algorithms focus on the HFL domain~\cite{kairouz2019advances,wang2021field} while VFL and FTL introduce additional federated optimization challenges that require more complex and expensive training protocols. 
Vertical federated learning requires an aggregation of the different features across learners and privacy-preserving computation of training loss and gradients across learners with~\cite{yang2019federated} or without~\cite{yang2019parallel,wu2020privacy} a third-party coordinator.
The most critical step in this domain is record linkage at the start of federated training through privacy-preserving entity resolution techniques~\cite{hardy2017private,xu2021fedv}. 
Federated transfer learning needs to learn common representations from the diverse learners' feature spaces and obtain predictions through one-side features~\cite{yang2019federated}.
Depending on the minimization domain, FTL methods can be further categorized into instance-, feature- or parameter-based~\cite{yin2021comprehensive}. In this work, we focus on the data integration and imputation challenges that arise in the Horizontal Federated Learning domains, but we hope our proposed solutions to spur further research into the VFL and FTL domains as well.

\section{\SystemDesignLong}

\label{sec:FLDesign}

\begin{figure*}[htbp]
  \centering
  \includegraphics[width=\linewidth]{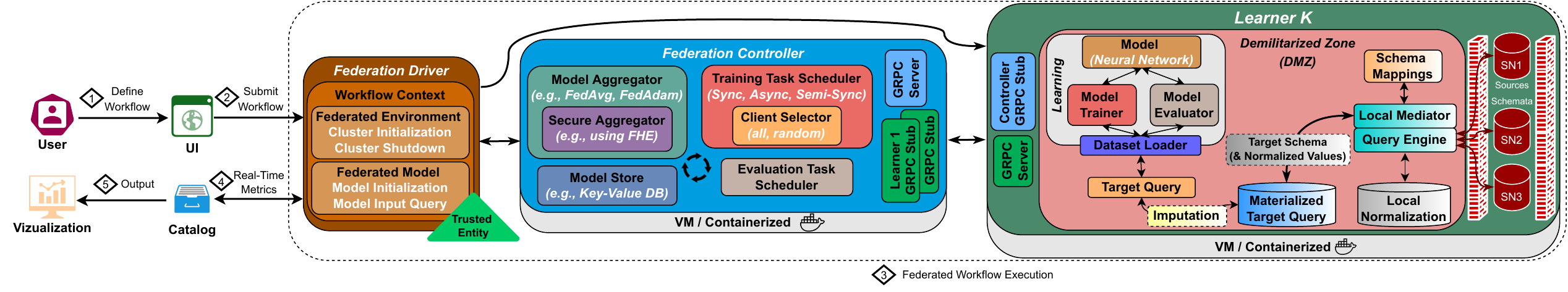}
  \caption{\SystemDesignLong~Architecture and Internal Components.}
  \label{fig:FLDMS_Design}
\end{figure*}

A scalable federated learning solution should adhere to the architectural principles of modularity, extensibility, and  configurability~\cite{beutel2020flower}.  \textit{Modularity} refers to the development of functionally independent services (micro-services) that allow a more fine control of system components' interoperability. \textit{Extensibility} refers to the functional interface expansion of each service.  \textit{Configurability} refers to the ease of deployment of new federated models and procedures. 
Following these principles, we propose a \SystemDesignLong~(\SystemDesignShort)~architecture (Figure~\ref{fig:FLDMS_Design}). 
We describe the core functions of the architecture in service of learning, data harmonization, and data imputation.

\subsection{Federated Learning Programming Model}

Following the successful programming model of Apache Spark~\cite{zaharia2010spark}, the Federation Controller operates as the cluster manager of the federation, Learners as the computing nodes, and the Driver as the entry point of the federation launching various operations in parallel.

\textbf{Federation Controller.} The controller orchestrates the distributed training of the federated model across learners. It comprises four main components. First, the Model Aggregator mixes the local models of the learners to construct a new global model. Second, the Training Task Scheduler manages learners' participation and synchronization points and delegates local training tasks.
The Model Store saves the local models and the contribution value of each learner in the federation to improve the efficiency of model aggregation over multiple training protocols~\cite{stripelis2022semi}. 
The Model Store component can be materialized through an in-memory or on-disk key-value store, depending on the number of learners and the size of the models (key: learner id, value: model and its contribution $p_k$).
The controller may also operate within an encrypted environment, in which case it needs to store encrypted local models and the global model aggregation function needs to be computed with homomorphic operations, such as the commonly-used weighted average methods~\cite{kaissis2020secure,zhang2020batchcrypt,stripelis2021secure}. Finally, the Evaluation Task Scheduler is responsible to dispatch the evaluation tasks to the learners and collect the associated metrics.

\textbf{Learner.} Every data silo acts as an independent learning entity that receives the global model and trains the model on its privately held local dataset through its Model Trainer component. A learner can also to support a Model Evaluator component to evaluate incoming models on its local training, validation and/or test datasets. Such evaluation can provide a score to weigh the models on actual learning performance~\cite{stripelis2020accelerating}. The Dataset Loader feeds \textit{harmonized} data to the training and evaluation components with the appropriate format. Section~\ref{sec:HarmonizVision} discuss the harmonization and imputation process in detail.

\par\textbf{Driver.} The Driver defines the high-level control flow of the federated application. Its main tasks are to initialize the Federation Controller and Learner services, define and initialize the neural network architecture (with a random or a pretrained model). 
The driver also collects real-time metadata associated with the federated training process and stores them inside the Catalog for further bookkeeping.
In our design, we consider the driver to be an independent trusted entity that can generate the key pairs of the encryption scheme.

\begin{figure}[htpb]
  \centering
  \includegraphics[width=0.8\linewidth]{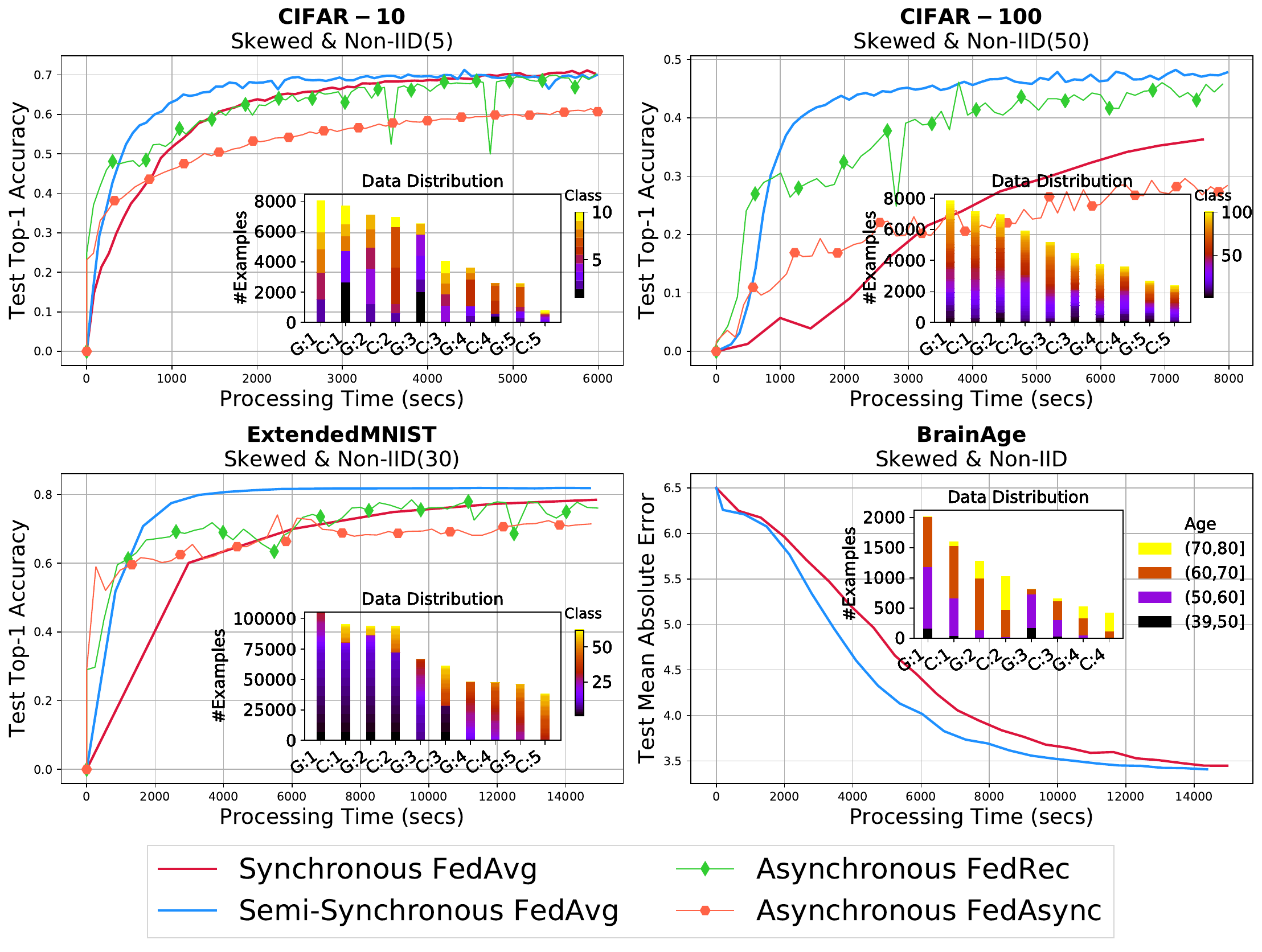}
  \caption{Federated Models Convergence in \metis.}
  \label{fig:metispoliciesdemo}
\end{figure}

\textbf{Evaluation.} An evaluation of the presented Federated Learning Programming Model is shown in~Figure~\ref{fig:metispoliciesdemo}. The figure shows the convergence of federated learning training policies for different communication protocols and model aggregation functions (FedAvg, FedRec, FedAsync)
on standard benchmarks and a neuroimaging domain using the \metis\ system~\cite{stripelis2022semi}.
For this evaluation, we consider a horizontal data partitioning scheme with all datasets conforming to the same schema, and training performed over complete data records with no missing values.
CIFAR is a benchmark for object detection in images (with 10 or 100 object classes).
ExtendedMNIST~\cite{caldas2018leaf} is a benchmark for recognition of handwritten letters and digits (we use ExtendedMNST ByClass with 62 unbalanced classes).
BrainAge~\cite{stripelis2021:isbi} is a neuroimaing task for predicting the age of a human brain from a structural MRI scan. The difference between the predicted and chronological age value is a biomarker of brain pathologies.
For CIFAR and ExtednedMNIST, we consider a computationally heterogeneous federation (5 CPUs and 5 GPUs) and for BrainAge a computationally homogeneous federation (8 GPUs). In all environments, the training data is non-IID across learners and learners hold different amounts of training samples (rightly skewed assignment; see plots' insets). The SemiSync policy has faster convergence, particularly in heterogeneous data and computational environments~\cite{stripelis2022semi}.

\subsection{Data Harmonization \& Imputation}
\label{sec:HarmonizVision}

All recently proposed Federated Learning systems~\cite{li2021survey} assume that the local training dataset of every silo conforms to the same data specifications. Such a scenario is not always true in real-world settings. Each silo is an independent entity and it is therefore natural to have its own unique data specifications. For example, in an international federation of hospitals, each institution may adhere to data specifications unique to the geographical region it operates on~\cite{louie2007data,cruz2007reviewing}.
Creating a consensus data model that can harmonize the nuances of such regional data specifications is a not an easy task, but it is critical for meaningful data analysis. Therefore, we propose to incorporate a data harmonization and integration component as a core feature of our architecture.

\textbf{Source Modeling/Schema Mapping}. 
The federated machine learning model needs a harmonized input across all participating sites. Although this could be accomplished by ad-hoc ETL pipelines at each site, such pipelines introduce maintenance and extensibility challenges.
To mitigate this, we advocate for a more principled declarative approach based on formal schema mappings following the vast work in data integration~\cite{halevy2001,fagin2005,gottlob2014s,doan2012,dong2015,xiao2018s}. 
First, we define a common schema (aka global, domain, mediated, or target schema) that represents an agreed-upon view of the application domain for the purposes of the federation. 
No more, no less; it does not necessarily need to provide the “perfect” model for the domain, but it needs to provide sufficient details to support the expected queries and analysis, and a reasonable expectation of extensibility as new sources are added.
Second, we define declarative schema mappings to translate the data from the sources into the common schema. These mappings are existential formulas of the form: 
{%
$\forall\vec{x},\vec{y}\
\phi_S(\vec{x},\vec{y})$  $\rightarrow \exists \vec{z}\
\psi_G(\vec{x},\vec{z})$}, where  $\phi_S$ and $\psi_G$ are conjunctions of predicates from the source and global (common) schemas, respectively, and $\vec{x}, \vec{y}, \vec{z}$ are tuples of variables. These mappings can be used for virtual data integration using query rewriting \cite{doan2012} %
or for data warehousing/data exchange \cite{fagin2005}. Complex constraints can be enforced on the target schema (i.e., being an ontology) and corresponding query answering methods exist \cite{gottlob2014s,xiao2018s}.
Declarative mappings have the advantage of being easier to generate, maintain, and provide opportunities for automatic learning and optimization (e.g.,~\cite{knoblock2012:eswc}). Figure~\ref{fig:schema_rules} shows an example of a global schema, schema mapping rules, and how queries on the domain schema support multiple learning tasks.

\textbf{Entity Linkage/Data Normalization}. It is also important to recognize when objects from different sources correspond to the same entity in the real world. For example, a patient may have interacted with several doctors, hospitals, testing facilities, pharmacies, etc., each of which may have created different records of these interactions. The data integration system must recognize that all theses records refer to the same patient, and link them into a complete medical history for the patient. When we deal with complex structured objects, such as patients, this problem is called entity or record linkage \cite{felligi1969,naumann2010,dong2015}. A simpler version of the problem also occurs with atomic values; different sources may use different strings to refer to the same value. For example, in a radiation oncology domain, one source may code an anatomical structure with a value “LTemp lobe,” while another uses the value “LTemporal.” To provide clear semantics for analysis, we need to map these two values to a normalized value such as "Left Temporal Lobe" (UBERON:0002808).

Figure~\ref{fig:FLDMS_Design} shows the detailed data harmonization components in our architecture. 
Each learner has an instance of a local mediator~\cite{doan2012,wiederhold92} with access to the schema mappings from its local source/s to the global schema. 
We envision that the federation will tackle different learning problems, at different times, over the same common view of the data that the mediator produces. Each learning problem would require a different neural network with different input. 
Thus, we obtain the required input data for each problem through a \textit{query} over the common schema, as opposed to ad-hoc ETL processes. The local data is never changed; however, our system can answer such queries using the schema mappings (and target schema constraints, if any) through query rewriting and data exchange techniques~\cite{halevy2001,fagin2005,gottlob2014s,doan2012,dong2015,xiao2018s}.
For data normalization, in simple cases we can use a local database with mappings between the values used in each source and normalized values, or use functional predicates that compute a similarity function between the source and the global values in more complex cases. These normalization relations can easily be added to the schema mappings  (Figure~\ref{fig:schema_rules}) and the query answering procedure as interpreted predicates, e.g., \cite{ambite2015}. 
The target query, which computes the input to the neural network, is materialized, so that the model trainer can efficiently operate over it.

Entity linkage across learners is more complex. So far we have discussed horizontal federated learning, where sites have similar id, feature and label space that is needed as input to the learning algorithm, possibly with imputed values. However, the required input data (i.e., data of a single learning example) may be distributed across several sites, so called vertical federated learning, where true private cross-source record linkage is needed~\cite{yang2019federated,yang2019parallel,wu2020privacy,hardy2017private,xu2021fedv}.

\textbf{Data Imputation.} After data harmonization, the machine learning model input is uniform and meaningful, since it is the output of a query over the common schema and values haven been normalized. However, real sources often have missing values, either missing at random or systematically.
One option is to remove rows or columns with missing values, but that diminishes the utility of the data and the quality of the learned models. It is generally preferable to \textit{impute} the missing values, that is, to find the most likely value for that given attribute and example~\cite{yoon2018gain,van2011mice}. Models learned with imputed values can lead to better performance~\cite{bertsimas2017predictive}.
In the context of federated learning, participating sources may have limited information, and statistically diverse data distributions, and therefore their local records may not be used to impute missing values/attributes. In these learning settings, an imputation function can be learned at the federation level. By training such a federated imputation function we can leverage the information from all sources, improve data quality and provide better data distribution coverage.

Data imputation interacts with formal query rewriting methods in an interesting way that opens new avenues for research. Since formal schema mappings have existential variables in the consequent, the query rewriting process may generate null values (skolems) in the answers to a query. Tuples with such null values are discarded, since they are not \textit{certain answers} (i.e., true in all possible worlds) \cite{doan2012,fagin2005}. However, for the purpose of learning, such null values can be imputed probabilistically. Therefore, we advocate to modify query answering algorithms to preserve null values and incorporate imputation procedures. 
Interestingly, the target query may need to retrieve attributes beyond those required by the input of the machine learning algorithm in order to improve the quality of the imputation.

\begin{figure*}[htbp]
\centering
\fbox{ \parbox{\linewidth}{
\begin{tabbing}
\textbf{Global schema} \\
clinical(id, visit, age, moca, dx) aa \= \kill
\textsf{subject(id, sex, re)  }               \> \# demographics, re = race/etchnicity \\
\textsf{clinical(id, visit, age, moca, dx)}   \> \# clinical data, visit = date of the assessment/dx,  dx should be icd10 codes \\
\textsf{imaging(id, visit, type, image) }     \> \# medical imaging of different types \\ 
\\
\textbf{Schema mappings} \\
impute\_f1(sex, age, re, mmse, moca\_imp, dx\_imp) aaaaa \= \kill
\textsf{s1(id, dob, sex, re, visit, mmse, dx, mri) $\wedge$ } \>  \# s1 only has MRIs, dx has missing values \\
\textsf{minus(dob, visit, age)  $\wedge$  }  \>    \# compute age at assessment as date of birth minus visit date \\
\textsf{impute\_f1(sex, age, re, mmse, moca\_imp, dx\_imp) } \>  \# imputation of MoCA (full column) and missing values of dx \\
\textsf{$\rightarrow$ subject(id, sex, re) $\wedge$ clinical(id, visit, age, moca\_imp, dx\_imp)
$\wedge$ imaging(id, visit, "MRI", mri) } \\
\\
s2\_scan(id, visit\_scan, age\_scan, scan\_type, scan) scan\_type = "MRI" aaaa aaaa aaaa aaaa \= \kill
\textsf{s2\_dem(id, sex, re) $\wedge$ }  \\
\textsf{s2\_image(id, visit\_image, age\_image, image\_type, scan) $\wedge$ image\_type = "MRI" $\wedge$ } \> \# only interesred in MRIs\\
\textsf{s2\_dx(id, visit\_dx, age\_dx, dx) $\wedge$ dx in ["CT", "MCI", "AD"] $\wedge$ } \> \# and on Alzhemer's diagnoses \\
\textsf{normalize(dx, icd10) $\wedge$  }    \>                  \# normalize diagnostic codes \\
\textsf{impute\_f2(sex, age\_dx, re, icd10, moca\_imp)  } \>   \# imputation of MoCA values, \\
\textsf{$\rightarrow$ subject(id, sex, re) $\wedge$ clinical(id, visit\_dx, age\_dx, moca\_imp, icd10) $\wedge$ 
imaging(id, visit\_image, "MRI", image) }\\
\\
\textbf{Alzheimer's prediction query} \\
q( \= \kill
\textsf{q(sex, re, age, mri, dx) $\leftarrow$ subject(id, sex, re) $\wedge$ imaging(id, visit1, "MRI", mri)  $\wedge$ } \\
\> \textsf{clinical(id, visit2, age1, moca, dx) $\wedge$ ( $|visit1 - visit2| < 60$ ) } \\
\\
\textbf{Cognitive decline query} \\
q( \= \kill
\textsf{q(sex, re, mri1, diff\_age, diff\_moca) $\leftarrow$ subject(id, sex, re) 
$\wedge$ imaging(id, visit1, "MRI", mri1)  $\wedge$ } \\
\> \textsf{clinical(id, visit1, age1, moca1, dx1)  $\wedge$ clinical(id, visit2, age2, moca2, dx2)  
$\wedge$  visit2 $>$ visit1 } \\
\> \textsf{minus(age1, age2, diff\_age) $\wedge$ minus(moca1, moca2, diff\_moca) }
 \end{tabbing}
}}
    \captionsetup[justification=center]{}
    \caption{Global Schema and Schema Mapping Rules.}
    \label{fig:schema_rules}
\end{figure*}

\textbf{Example}. Figure~\ref{fig:schema_rules} shows a notional example of horizontal federated learning and integration (FLINT) on sources with medical data. The federation designers define a (harmonized) global schema with 3 relations: \textsf{subject}, which models subject demographics; \textsf{clinical}, which models clinical assessments and diagnoses, and \textsf{imaging}, which models different types of medical imaging. The federation expects normalized icd10 codes for diagnoses, and standardizes on the Montreal Cognitive Assessment (MoCA) as a measure for dementia. 
There are two sources:  \textsf{s1}, which represents a clinic specializing in the treatment of Alzheimer's Disease that captures magnetic resonance imaging (MRI) and administers a Mini-Mental State Examination (MMSE) for each patient, both on a single visit; and \textsf{s2}, which represents a hospital that treats a wider variety of diseases. 
The two sources are mapped to the global schema using formal schema mappings that include both data transformation and imputation functional predicates. 
The first mapping uses a simple functional predicate to compute the age at assessment from the difference of the patient date of birth and the visit date, as well as an imputation procedure that imputes both the MoCA score and possibly missing diagnosis codes from the MMSE score, age, sex, race/ethnicity, and existing diagnosis. Since \textsf{s1} contains only MRIs, this is the recorded type of the resulting imaging in the harmonized schema. 
The second mapping joins 3 tables from source \textsf{s2} comprising demographics, imaging, and diagnoses. Assume the federation is only interested in neuroimaging of Alzheimer's Disease, so it chooses to populate the global schema only with MRI scans and relevant diagnoses (AD: Alzheimer's Disease, MCI: mild cognitive impairment, and controls). An interpreted predicate maps the source's diagnoses to appropriate ICD10 codes. Finally, a second imputation function imputes the MoCA values from sex, age, race/ethnicity and diagnosis. Note that since the MoCA score is not produced by the source, but it is required by the global schema predicate \textsf{clinical}, it would have been represented by an skolem function in a traditional data integration system, and query tuples with such skolem would have been removed. Here, we impute the MoCA score, so no tuples are lost. 

The global schema, through the schema mappings, enables a variety of queries that support different learning tasks within the federation. Figure~\ref{fig:schema_rules} shows two such queries. The first computes the training data for a {\em classification} learning task that predicts an AD status (AD/MCI/CT) diagnosis based on the MRI, sex, age, and race/ethnicity of a subject. The second query computes the training data for a {\em regression} learning task to predict cognitive decline based on an MRI at an initial time point and the ages and cognitive assessment values at two timepoints.

\textbf{Privacy.} Since data privacy is critical in federated learning, query rewriting and data normalization need to be performed locally at each site and therefore the schema mappings and the normalization tables need to be kept local at each source and only the global model schema is shared across sources. Similarly, record linkage can be done in a privacy-preserving manner~\cite{scannapieco2007,liang2004} but federated training becomes significantly more complex, which is an active area of research~\cite{cha2021,hardy2017private,xu2021fedv,yang2019federated}.

\section{Discussion}
We presented an architecture for a \SystemDesignLong~(\SystemDesignShort) platform for distributed training across a federation of data silos, including data integration and imputation components, which are critical for meaningful analysis. 
We advocated using principled data harmonization methods, leveraging the vast literature on data integration~\cite{halevy2001,fagin2005,gottlob2014s,doan2012,dong2015,xiao2018s}. 
Specifically, we proposed to model the application domain through a target schema and formal schema mappings, and to execute target queries to provide the input data for the federated learning model.
Since the purpose of data integration is analysis, we propose new research directions for query answering techniques to incorporate statistical imputation (instead of discarding answers with labeled nulls). 
We plan to release \metis\ as an open-source prototype \SystemDesignShort~to stimulate further research on the interaction of databases and machine learning.

\section{Acknowledgments}
\label{sec:Acknowledgements}
This research was supported in part by the Defense Advanced Research Projects Agency (DARPA) under contract HR0011\-2090104, and in part by the National Institutes of Health (NIH) under grant R01DA053028.  

\bibliography{main.bib,db-kr,ambite}

\end{document}